\documentclass[preprints,article,accept,moreauthors,pdftex,10pt,a4paper]{Definitions/mdpi}

\usepackage{gensymb}

\usepackage{CJKutf8}
 
\usepackage{longtable}

\usepackage{soul,color}

\usepackage{amsthm}
\usepackage{relsize}

\usepackage{adjustbox}
\usepackage{mathtools}
\usepackage{hhline}
\usepackage{array}
\usepackage{makecell}
\usepackage{stackengine}
\setstackEOL{\cr}

\usepackage{listings}
\usepackage{color}

\definecolor{dkgreen}{rgb}{0,0.6,0}
\definecolor{gray}{rgb}{0.5,0.5,0.5}
\definecolor{mauve}{rgb}{0.58,0,0.82}

\definecolor{codegreen}{rgb}{0,0.6,0}
\definecolor{codegray}{rgb}{0.5,0.5,0.5}
\definecolor{codepurple}{rgb}{0.58,0,0.82}
\definecolor{backcolour}{rgb}{0.95,0.95,0.92}
 
\lstdefinestyle{mystyle}{
    language=Python,
    aboveskip=5mm,
    belowskip=5mm, 
    numberstyle=\tiny\color{codegray},
    stringstyle=\color{codepurple},
    commentstyle=\color{dkgreen},
    keywordstyle=\color{blue},
    frame = lines,
    basicstyle=\tt\4, 
    breakatwhitespace=true,         
    breaklines=true,           
    basicstyle=\selectfont\rmfamily,
    captionpos=t,                    
    keepspaces=true,                 
    numbers=left,                    
    numbersep=2pt,                  
    showspaces=false,                
    showstringspaces=false,
    stepnumber=1, 
    showtabs=false,                  
    tabsize=3,
    moredelim=[is][\underbar]{_}{_} 
}
 
\usepackage{fullpage}
\usepackage{float}
\usepackage{times}
\usepackage{fancyhdr,graphicx,amsmath,amssymb}
\usepackage[linesnumbered,ruled,vlined]{algorithm2e} 

\SetKwProg{Fn}{Function}{}{} 

\firstpage{1} 
\pubvolume{xx}
\issuenum{1}
\articlenumber{5}
\pubyear{2018}
\copyrightyear{2018}
\history{Received: date; Accepted: date; Published: date}
\Title{A Study of Machine Learning Models in Predicting the Intention of Adolescents to Smoke Cigarettes}

\Author{Seung Joon Nam $^{1}$, Han Min Kim $^{2}$, Thomas Kang $^{3}$, Cheol Young Park$^{4,}$*}

\AuthorNames{Firstname Lastname, Firstname Lastname and Firstname Lastname}

\address{%
$^{1}$ \quad Robinson Secondary School, USA; seungjoonnam@gmail.com\\
$^{2}$ \quad Centreville High School, USA; guppy1@naver.com\\
$^{3}$ \quad West Springfield High School, USA; tk3477@gmail.com\\
$^{4}$ \quad BAIES, Bayesian AI Lab, USA; cparkf@gmu.edu}

\corres{Correspondence: cparkf@gmu.edu}

\abstract{The use of electronic cigarette (e-cigarette) is increasing among adolescents. This is problematic since consuming nicotine at an early age can cause harmful effects in developing teenager's brain and health. Additionally, the use of e-cigarette has a possibility of leading to the use of cigarettes, which is more severe. There were many researches about e-cigarette and cigarette that mostly focused on finding and analyzing causes of smoking using conventional statistics. However, there is a lack of research on developing prediction models, which is more applicable to anti-smoking campaign, about e-cigarette and cigarette. In this paper, we research the prediction models that can be used to predict an individual e-cigarette user's (including non-e-cigarette users) intention to smoke cigarettes, so that one can be early informed about the risk of going down the path of smoking cigarettes. To construct the prediction models, five machine learning (ML) algorithms are exploited and tested for their accuracy in predicting the intention to smoke cigarettes among never smokers using data from the 2018 National Youth Tobacco Survey (NYTS). In our investigation, the Gradient Boosting Classifier, one of the prediction models, shows the highest accuracy out of all the other models. Also, with the best prediction model, we made a public website that enables users to input information to predict their intentions of smoking cigarettes.}


\begin{document}


\section{Introduction}

The number of high school students who use e-cigarette increased by 78\% resulting in 3.05 million and the number of middle schoolers who use e-cigarette increased by 48\% bringing the total up to 570,000 between 2017-2018 \cite{2018NYTSData}. This brings the total, according to 2018 National Youth Tobacco Survey (NYTS), to about 3.6 million middle and high school students using e-cigarette \cite{2018NYTSData}. The problems with youths using e-cigarette is that exposure to nicotine at an early age can cause addiction and harm the developing brain \cite{ECIGFACT}. Also, Bunnell et al. \cite{bunnell2015intentions} found that e-cigarette use has associations with smoking cigarettes. Due to the fact that e-cigarette use is relatively a recent issue, there are only a small number of research done regarding the use of e-cigarette for adolescents \cite{dawkins2013vaping, palazzolo2013electronic, miech2017cigarette, raloff2015dangers, dutra2014electronic, kosmider2014carbonyl, littlefield2015electronic, pepper2017risk, pepper2017risk,huey2017escape, schneider2015vaping,mccabe2017associations,penzes2018bidirectional,arnold2014vaping,miech2017kids,schripp2013does,westling2017electronic,mccabe2017smoking,tsai2018reasons}. Dawkins et al. \cite{dawkins2013vaping} discusses how vaping is used to decrease the dependence on cigarettes and that e-cigarettes are better than nicotine replacement therapy. Some examples of side effects of vaping were throat irritation and mouth irritation but only a small percent of people had these effects. In another research, Palazzolo \cite{palazzolo2013electronic} states that people are not sure of the harmful results of vaping due to lack of empirical data, but there were some cases of people negatively affected. Also, McCabe et al. \cite{mccabe2017smoking} states that students who smoke e-cigarettes early on tend to smoke as they get older, so it is a gateway drug to other worse drugs. Chapman \cite{chapman2014cigarettes} states that vaping could have a positive impact on people to transition from smoking cigarettes to vaping, which helps people to move away from smoking. However, a negative impact for non-smokers would be to move to vaping, which could lead to other more harmful drugs. Another research carried by Miech \cite{miech2017cigarette} states that people who vaped were 4 times more likely to smoke cigarettes, and vaping does not predict the decrease in smoking. In the research by Dutra \cite{dutra2014electronic}, it is stated that adolescents who have vaped before have a higher chance of smoking cigarettes. Also, most of the vapers smoked cigarettes as well. According to the research by \cite{huey2017escape}, more than 250,000 teens who never smoked a regular cigarette have vaped and those who did are twice more likely to smoke regular cigarette in the future.

It is evident from the multiple research papers that vaping leads to smoking cigarettes later on. However, there is a methodological limitation among these researches. Most of the research utilized statistical methods to analyze the relationship between causes and effects for e-cigarette and/or smoking habit. For example, one research\cite{bunnell2015intentions} discusses the association between e-cigarette use and smoking intention among US youths who have never smoked a traditional cigarette. The article implemented chi-squared tests, multivariate logistic regression, and other models to analyze the data. Although analyzing data using logistic regression model was appropriate for their case of study, it may not be the most practical in terms of anti-smoking campaign, preventing smoking habit. Using prediction models developed from machine learning (ML) algorithms would be more useful, because it can predict whether a person will have the intention to smoke cigarette or not depending on his or her information about e-cigarette use as well as race, ethnicity, gender, and the environment. Additionally, a prediction can directly help individuals stay away from the path of smoking cigarettes, which is more helpful than simply analyzing data. 

In this paper, we use the NYTS results from 2018 and construct multiple prediction models (e.g.,Gradient Boosting Classifier, and Decision Tree Classifier) that can predict whether a person will have an intention to smoke cigarettes or not. After data analysis, Gradient Boosting Classifier, one of the prediction models, had the highest accuracy of 93\% out of all the models tested. We divide the data into two sets: never-smokers and smokers of cigarette. The group of never-smokers were analyzed to find the best fitting model to predict the intention to smoke cigarettes for both e-cigarette smokers and non-e-cigarette smokers. In addition, we create a website involving the Gradient Boosting Classifier model in order to allow the public to input factors (e.g., sex, race, and age) and receive a prediction of whether or not they will have a high intention to smoke cigarettes or not. This will give the general public more awareness of their position as to whether or not they will smoke cigarettes and possibly steer away from the path of smoking cigarettes.

Consequently, there are two contributions in this paper: (i) found the best-fitting model to predict smoking intention from the NYTS data, and (ii) create a website to help students, especially e-cigarette smokers, be able to prevent e-cigarette use due to possible chance of smoking cigarette.

This paper is organized as following: Section 2 introduces background of ML techniques; Section 3 presents analysis methods; Section 4 explains prediction system for electronic cigarette usage based on smoking intention; Section 5 includes the conclusion and future study. 

\section{Background of Machine Learning Techniques}
In this section, we introduce the CRISP-DM (Cross-Industry Standard Process for Data Mining) process model \cite{wirth2000crisp}. The CRISP-DM process model is a comprehensive and simple method for data mining and analysis. Throughout Section 2.1, we describe the six steps of the CRISP-DM process model. In addition, five machine learning algorithms used in this research are introduced in Section 2.2.

\subsection{CRISP-DM}
\begin{figure}[H]
\centering
\includegraphics[width=\textwidth]{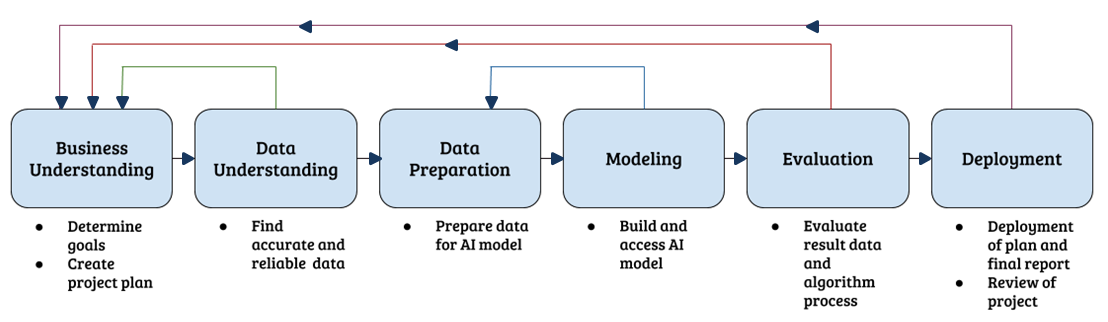}
\caption{Process of CRISP-DM}
\label{fig:crisp}
\end{figure}

\unskip
CRISP-DM \cite{wirth2000crisp} (see Figure \ref{fig:crisp}) is a cross-industry standard process for data mining and analysis. CRISP-DM helps plan out a solid method of running a data mining project and directs at making data mining and analysis projects more efficient, cheaper, and faster. 
CRISP-DM is comprised of six steps: (1) business understanding, (2) data understanding, (3) data preparation, (4) modeling, (5) evaluation, and (6) deployment.  

\subsubsection{Business Understanding}
\unskip

\textit{Business Understanding} is introducing the objectives of the project and understanding what the goals will be. Knowing the goals and the objectives, a data mining problem is formed and a plan to achieve these goals is made. Figure \ref{fig:crisp} is the CRISP-DM process and the first box covers the business understanding step. 

\subsubsection{Data Understanding}
\unskip

\textit{Data Understanding} includes finding an initial data set, familiarizing with the contents in the data set, and making observations from the data set in order to create an hypothesis. The initial data set must be reliable and accurate, which means that it is not outdated and has correlation within itself. The second box from Figure \ref{fig:crisp} indicates the data understanding step.  

\subsubsection{Data Preparation}
\unskip

\textit{Data Preparation} is the process of converting the initial, or raw, data set into the final data set. This step could alter the data to become more applicable to achieve the goal. The third box from Figure \ref{fig:crisp} covers the data preparation step. 

\subsubsection{Modeling}
\unskip

In \textit{Modeling}, artificial intelligence models are created using various machine learning algorithms. The data set prepared from the data preparation step are applied to this modeling step. 

\subsubsection{Evaluation}
\unskip

\textit{Evaluation} is reviewing performance (e.g., accuracy) of models and deciding a best model. This is the process where the models are improved based a goal performance. Success criteria can be based on speed of algorithm, memory usage, or prediction accuracy. According to analysis purpose, different performance metrics are used. For example, for regression problem, the $R^2$ score can be used. For classification problem, an accuracy ($Acc$) of Equation \ref{eq:acc}, the sum of correct classification divided by the total number of classifications, can be used. Since this paper uses classification, some criteria for classification are introduced in the following. 

\begin{equation}
\label{eq:acc}
  Acc =  \frac{The\ number\ of\ correct\ classification}{The\ total\ number\ of\ classification} = \frac{\sum_{i=1}^{N}x_{ij}}{\sum_{i=1}^{N}\sum_{j=1}^{N}x_{ij}},
\end{equation} 
where $N$ denotes the number of class and $x_{ij}$ denotes the total number of the case in which values of $i$-th prediction and $j$-th observation are identical.

In this study, we used the four-performance metrics Acc (Equation \ref{eq:acc}), Precision (Equation \ref{eq:prec}), Recall (Equation \ref{eq:recall}), and F1-score (Equation \ref{eq:f1}). These metrics can be easily calculated by using the following four indicators.

\begin{itemize}
  \item True Positive (TP): The amount of the observed positive values which were correctly predicted
  \item False Positive (FP): The amount of the observed positive values which were wrongly predicted
  \item False Negative (FN): The amount of the observed negative values which were wrongly predicted
  \item True Negative (TN): The amount of the observed negative values which were correctly predicted
\end{itemize}

These four indicators can be used to define the equations of Precision and Recall as shown.
 
\begin{equation}
\label{eq:prec}
  Precision =  \frac{TP}{TP+FP}
\end{equation} 

\begin{equation}
\label{eq:recall}
  Recall =  \frac{TP}{TP+FN}
\end{equation} 

Precision is commonly used to measure the influence of False Positive, while Recall is used to measure the influence of False Negative. F1-score is defined as the weighted average of Precision and Recall.

\begin{equation}
\label{eq:f1}
  F1 \text- Score =  \frac{2*(Precision*Recall )}{Precision+Recall}
\end{equation}

Precision, Recall, and F1-score have a score of one when the prediction is perfect. For the total prediction failure, they yield a score of zero.

\subsubsection{Deployment}
\unskip

\textit{Deployment} is the process of organizing the information gained, such as the model, so that it is understandable for the model user. This step is carried out by the user rather than the analyst, so it is essential for the user to understand how to use the models. The result of this step is a final report. The last step in Figure \ref{fig:crisp} represents the deployment step.  
 
\subsection{Machine Learning Models }
\unskip
In this subsection, we introduce machine learning models that this paper utilizes. Since \textit{Linear Regression} model is a basic model in machine learning, we start with this model. Then, other models (i.e., \textit{Logistic Regression Classifier}, \textit{Gaussian Naive Bayes Classifier}, \textit{Decision Tree Classifier}, \textit{Random Forest Classifier}, and \textit{Gradient Boosting Classifier}) are introduced. 

\subsubsection{Linear Regression Model}
\unskip

Linear Regression model represents a linear trend of data and is used to predict a \textit{continuous response} $a(x)$ using \textit{predictor variables} $x$. We define \textit{Linear Regression} model formally. 

\begin{equation}
\label{eq:RM}
\begin{split}
& a(x) = w^Tx, 
\end{split}
\end{equation}  
where $w$ denotes weights (or coefficients) \{$w_1$, $w_2$, ..., $w_d$\}, $x$ denotes features \{$x_1$, $x_2$, ..., $x_d$\}, and $d$ denotes the number of weights.

In these settings, machine learning of Linear Regression model is to estimate better weights (or parameters) $w$ which fits data. A simple approach is a least squares approach in which the parameters are found by minimizing a \textit{residual sum of squares} (RSS).  

\begin{equation}
\label{eq:RRS}
\begin{split}
& RRS = \sum_{i=1}^n e^2_{(i)},  
\end{split}
\end{equation}  
where $e_{(i)} = y_{(i)}-a(x_{(i)})$ denotes an \textit{i}-th \textit{residual error} for \textit{i}-th data (or row) consisting of a pair of $\{x_{(i)}, y_{(i)}\}$ and $n$ denotes the number of data.  

Using RRS (Equation \ref{eq:RRS}), a gradient descent algorithm can be applied to find the best weights $w$ fitting training data. The process of the gradient descent algorithm follows: (1) Initial weights are randomly selected, (2) The weights are updated according to 

\begin{equation}
\label{eq:gdalg}
\begin{split}
& w_i = w_{i-1} + r\frac{\partial}{\partial w_{i-1}}RRS,
\end{split}
\end{equation}  
where $w_{i-1}$ denotes a previous weight of $w_{i}$ and $r$ is a learning rate, (3) if $||w_{i}-w_{i-1} || < e$, where $e$ denotes a threshold for the convergence, then stop.

\subsubsection{Logistic Regression Classifier}
\unskip
Logistic Regression Classifier is an extension of the Linear Regression model. Especially, the response $a(x)$ in Logistic Regression is discrete (e.g., \textit{True} and \textit{False}). In other words, given predictor variables $x$, a discrete value $y = a(x)$ is predicted. A simple case of the Logistic Regression is a binary Logistic Regression model whose output is Boolean. The binary Logistic Regression model can be defined formally as shown Equation \ref{eq:LR}.

\begin{equation}
\label{eq:LR}
\begin{split}
& a(x) = \frac{1}{1+e^{-w^Tx}}, 
\end{split}
\end{equation}  
where $e^{(.)}$ is a natural exponential function and $w^Tx$ is identical to Equation \ref{eq:RM}.

For a general Logistic Regression, whose output is multiple, a softmax function can be applied to represent 
a probability distribution for multiple classes. Equation \ref{eq:GLR} shows the general Logistic Regression.

\begin{equation}
\label{eq:GLR}
\begin{split}
& a(x) = \Bigg(\frac{e^{w_{(1)}^Tx}}{\sum_{i=1}^{K}e^{w_{(i)}^Tx}},...,\frac{e^{w_{(K)}^Tx}}{\sum_{i=1}^{K}e^{w_{(i)}^Tx}}\Bigg), 
\end{split}
\end{equation}  
where $K$ denotes the number of classes.

For machine learning, the Logistic Regression model requires a loss function to measure similarity between the leaned model and data, like the residual sum of squares in the Linear Regression model. Equation \ref{eq:GLR_loss} shows the loss function of the general Logistic Regression.

\begin{equation}
\label{eq:GLR_loss}
\begin{split}
& Loss = -\mathlarger{\mathlarger{\sum}}_{k=1}^{K}[y=k]\log \frac{e^{w_{(k)}^Tx}}{\sum_{i=1}^{K}e^{w_{(i)}^Tx}}, 
\end{split}
\end{equation}  
where $[s]$ is a function return one $s$ is true, zero otherwise.

This loss function is used for the gradient descent algorithm (see Equation \ref{eq:gdalg}) to find best weights $w$ by substituting RSS in Equation \ref{eq:gdalg} with the Loss function.  
 
\subsubsection{Gaussian Naive Bayes Classifier}
\unskip

Naive Bayes (NB) Classifier \cite{maron1961automatic} is a classification method based on probability theory, by which one can classify class labels using (discrete or continuous) inputs. Basically, NB represents a joint distribution for input random variables ($X$) and a class random variables ($Y$) on the assumption of conditionally independence of $X$ given $Y=c$. NB can be written as Equation \ref{eq:NB}.

\begin{equation}
\label{eq:NB}
\begin{split}
& P(X, Y=c) = P(Y=c)\mathlarger{\mathlarger{\prod}}_{i=1}^{K}P(X_{i}|Y=c), 
\end{split}
\end{equation}  
where $K$ is the number of input random variables and $c$ denotes a class label in $Y$.

Gaussian Naive Bayes Classifier uses continuous inputs under the Gaussian (or Normal) distribution assumption. 
\begin{equation}
\label{eq:GNB}
\begin{split}
& P(X, Y=c) = P(Y= c)\mathlarger{\mathlarger{\prod}}_{i=1}^{K}N(X_{i}|\mu_{i, c}, \sigma_{i, c}), 
\end{split}
\end{equation}  
where $\mu_{i, c}$ and $\sigma_{i, c}$ denote the mean and standard deviation of the input $i$ for the class label $c$, respectively.

For machine learning of Gaussian Naive Bayes Classifier, Maximum Likelihood Estimation (MLE), Maximum a Posteriori Probability (MAP), and Bayesian approach can be used \cite{murphy2012machine}. 

\subsubsection{Decision Tree Classifier}
\unskip

Decision tree Classifier \cite{quinlan1986induction, morgan1963problems, safavian1991survey, magerman1995statistical} consists of a tree structure containing a set of hierarchical nodes (Root Node, Internal Nodes, and Leaf Nodes). The root node and the internal nodes represent features or variables, while the leaf nodes denote values of a target variable. The main challenge of machine learning for decision tree is to construct these nodes and their hierarchy in a decision tree, so that it can effectively classify classes using input data of predictor variables. The basic approach, called ID3 (Iterative Dichotomiser 3), to build a decision tree was introduced by \cite{quinlan1986induction}. For example, suppose that there is a target variable with two classes (Positive and Negative), the expected information for this can be written by $I(.,.)$. Note that the following equations are taken from \cite{quinlan1986induction}.

\begin{equation}
\label{eq:entropy}
\begin{split}
& I(p, n) = - \frac{p}{p+n} \log_{2} \frac{p}{p+n} - \frac{n}{p+n} \log_{2} \frac{n}{p+n},
\end{split}
\end{equation}  
where $p$ and $n$ denote the numbers of positive and negative cases, respectively. The expected information $E(.)$ for a parent node $A$ of the target variable can be derived as the weighted average.
\begin{equation}
\label{eq:entropy1}
\begin{split}
& E(A) = \sum^{v}_{i=1} \frac{p_i+n_i}{p+n} I(p_i, n_i),
\end{split}
\end{equation}  
where $v$ denotes the number of the parent node values and $I(p_i, n_i)$ denotes the expected information for the $i-$th value of the parent node. The information $gain(A)$ for the node $A$ can be obtained as the following equation. 
\begin{equation}
\label{eq:entropy2}
\begin{split}
& gain(A) = I(p, n) - E(A).
\end{split}
\end{equation}  

Thus, machine learning for decision tree  is to find a tree on maximizing the information $gain(.)$ for all the root and internal nodes

\subsubsection{Random Forest Classifier}
\unskip
A set of ML models can often have a better performance than the use of a simple ML model. Such integration of ML models is called an ensemble learning. Random Forest Classifier \cite{ho1995random, ho2002data} uses the ensemble learning by forming a set of decision trees and resulting in an output which are voted from each decision tree. Random Forest draw random samples from training data and learn a decision tree model from the sample data, so that it can have a set of decision trees (i.e., forest). After machine learning, in the prediction (or application) stage, the class voted by the majority of learned decision trees is chosen as the final result. The following shows an equation for such majority voting. 

\begin{equation}
\label{eq:RF}
\begin{split}
& \hat{y} = mode\{a_1(x), a_2(x), ..., a_n(x)\},  
\end{split}
\end{equation}  
where $a_i(x)$ is a single decision tree and the function $mode(.)$ yields the output as the class label that is the most frequent class among the set of classification results.
 
\subsubsection{Gradient Boosting Classifier}
\unskip
Gradient Boosting Classifier \cite{breiman1996arcing} uses an ensemble model consisting of a set of simple models (e.g., a decision tree stump, a tree containing only one root and its immediately connected leaf nodes). By adding such simple models, the result ensemble model can be sequentially improved and finally fitted to data. In other words, after applying a simple model, samples which are classified by it are reused to another simple model. And then this process is repeated until convergence (or achieving better predictive performance). Gradient Boosting Classifier is a generalized method of boosting (e.g., \cite{schapire1990strength, freund1996experiments}) by using gradient of a loss function.

\section{Analysis Methods}
In this section, we introduce the specific processes of our analysis and the results from the analysis.

\subsection{Business Understanding}
\unskip
In this paper, our goal has two folds of objectives. (1) The first goal is to determine the most accurate machine learning model between Linear Regression, Gaussian Naive Bayes, Decision Tree, Random Forest, and Gradient Boosting. (2) The second is to construct a public web that will allow adolescents to know whether they will have the intention to smoke or not. 

We utilize the data from NYTS for 2018 in order to construct ML models, and the models are analyzed to choose the most accurate one in predicting the intention for a non-smoker to smoke cigarettes. 

\subsection{Data Understanding}
\unskip


We found reliable data from the \textit{National Youth Tobacco Survey 2018} \cite{2018NYTS_DataSet}, which is a nation-wide survey in US for middle (grades 6-8) and high school (9-12) youth's tobacco-related beliefs, attitudes, behaviors, and exposure to pro and anti-tobacco influences. NYTS implements a three-stage cluster sampling design in order to get a nationally representative data for students from grades 6-12 in all 50 states and District of Columbia \cite{NYTS}. However, the data covers only the youth who are currently attending middle and high school, which means that the results would be inapplicable for youths who are not attending middle and high school.

There are a total of 88 questions in the survey (Table \ref{tb:quest88}), and we examined all of the questions and picked the questions that were necessary making the model to predict the whether a person has the intention to smoke or not. 
Table \ref{tbl:questions} describes the format of the questions and the answer choices as written on the questionnaire. For example, in the first row of Table \ref{tbl:questions}, Q1 represents question 1 and the next column describes what the actual question is, which is "How old are you?" The possible answer choices is displayed on the next column and they are ranged from ages 9-19 years old. 

\begin{table}[H]
\caption{Illustrative Example of Questions} 
\label{tbl:questions}
 
\centering 
\begin{tabular}{c c c}
\toprule 
\thead{\textbf{Question Number}} & 
\thead{\textbf{Question}} &
\thead{\textbf{Answers}}  \\
 
\midrule
Q1 & How old are you? & 9, 10, ..., 19 years old\\ 
Q2 & What is your sex? & Male/Female \\ 
Q3 & What grade are you in? & 6, 7, ..., 12, ungraded or other grade\\
Q4 & Are you Hispanic, Latino, Latina, or of
Spanish origin? & Select one or more \\
... & ... & ... \\ 
Q88 & Because of a physical...making decisions? & No/Yes \\ 

\midrule 
\end{tabular}
\end{table}  

\subsection{Data Preparation}
\unskip
Our original data was taken from the National Youth Tobacco Surveys 2018. This data set was compiled of questions represented as questions Q and answers represented by numbers (e.g., 1, 2, 3, and 4) and words (e.g., Yes and No). We filtered every question to prepare for machine learning. For example, any null in the answer meant that the question was not answered. We went through the process of replacing all the nulls with 0's, which represents the unanswered choices. 

The next process involved classifying related data. When we first found the data, it was a set of 20189 rows x 195 columns. As mentioned in Section 1, our goal is to construct a prediction model construct a public web. In order to achieve this goal, we needed to divide the data into two groups: never smoked cigarette users and ever smoked cigarette users. Then, all the rows containing individuals who have ever smoked in their lives were deleted, because we only need to analyze the youths who never smoked cigarettes. The purpose of the first split of the cigarette and non-cigarette users is to form a predictive model. Next, we extracted our target question and the questions pertaining to it. There are 88 questions within the survey we used, but not every question was related to our prediction. After examining all the survey questions, we chose specific questions to be used, since not all the questions were applicable to our goal and redundancies and indirect correlations were present. Out of 88 questions, 47 questions were used, and specifically questions 15, 16, 17, 43, 44, and 45 \textit{"do you think that you will try a cigarette soon?"}, \textit{"do you think you will smoke a cigarette in the next year?"}, \textit{"if one of your best best friends were to offer you a cigarette, would you smoke it?"}, \textit{"do you think that you will try smoking tobacco in a hookah or waterpipe soon?"}, \textit{"do you think you will smoke tobacco in a hookah or waterpipe in the next year?"}, \textit{"if one of your best friends were to offer you a hookah or waterpipe with tobacco, would you try it?"}, respectively, was our target questions, because they are all related to someone's intention to smoke a cigarette, no matter how small the intention. We decided that the answer choices for target questions (definitely yes, probably yes, probably no, definitely no) could be simplified to two answer choices by considering definitely yes and probably yes as "yes" and definitely no and probably no as "no". 

\subsection{Modeling}
\unskip
We used five ML algorithms to generate each ML model using a training data and evaluated the accuracy for each one of the models. The models include (1) Decision Tree Classifier, (2) Gaussian NB Classifier, (3) Logistic Regression Classifier, (4) Gradient Boosting Classifier, and (5) Random Forest Classifier. Out of the whole data set in Subsection 3.3, we assigned the training data set, which is 80 percent of the original data, and the test data set, which is 20 percent of the original data. The training data set is used to learn the prediction model while the test data set is used to test the learned model by evaluating the accuracy for each model. After examining the predicted accuracy, we can evaluate the model that showed the highest accuracy, which can be seen in Subsection 3.5.

\subsection{Evaluation}
\unskip

Table \ref{tbl:dtc} to Table \ref{tbl:gb} shows the results of the prediction accuracy that was taken from the five machine learning models. Note that the terms precision, recall, and F1-score that are used for the table are explained in Subsection 2.1.5. The first column for each table lists the answer choices for question 16 "Do you think you will smoke a cigarette in the next year?". This step is to see how the factors are related to our target question, Q16, and we used machine learning models to find which factor affects people to answer Q16 which is intention to smoke cigarettes. 

\begin{table}[H]
\caption{Decision Tree (DT)} 
\label{tbl:dtc}
\centering 
\begin{tabular}{c c c c |c}
\toprule   
\thead{\textbf{}} & 
\thead{\textbf{Precision}} &
\thead{\textbf{Recall}} &
\thead{\textbf{F1-Score}} &
\thead{\textbf{Cases}} \\
 
\midrule
Yes & 0.60 & 0.42 & 0.49 &  201 \\ 
\midrule
No &  0.94 & 0.97 & 0.96 & 2049   \\
\bottomrule
Macro Avg &  0.77 & 0.70 & 0.72 & 2250 \\
\bottomrule
Weighted Avg &  0.91 & 0.92 & 0.92 & 2250

\end{tabular}
\end{table}  
\begin{table}[H]
\caption{Gaussian Naive Bayes (NB)} 
\label{tbl:nb}
\centering 
\begin{tabular}{c c c c |c}
\toprule   
\thead{\textbf{}} & 
\thead{\textbf{Precision}} &
\thead{\textbf{Recall}} &
\thead{\textbf{F1-Score}} &
\thead{\textbf{Cases}} \\ 

\midrule
Yes &  0.25 & 0.77 & 0.38 &  201 \\ 
\midrule
No &   0.97 & 0.77 & 0.86 & 2049  \\
\bottomrule
Macro Avg &  0.61 & 0.77 & 0.62 & 2250 \\
\bottomrule
Weighted Avg &   0.91 & 0.77 & 0.82 & 2250

\end{tabular}
\end{table}  

\begin{table}[H]
\caption{Logistic Regression (LR)} 
\label{tbl:lg}
\centering 
\begin{tabular}{c c c c |c}
\toprule   
\thead{\textbf{}} & 
\thead{\textbf{Precision}} &
\thead{\textbf{Recall}} &
\thead{\textbf{F1-Score}} &
\thead{\textbf{Cases}} \\
  
\midrule
Yes &  0.66 & 0.30 & 0.41 &  201 \\ 
\midrule
No &  0.93 & 0.98 & 0.96 & 2049  \\
\bottomrule
Macro Avg &  0.80 & 0.64 & 0.69 & 2250 \\
\bottomrule
Weighted Avg &   0.91 & 0.92 & 0.91 & 2250

\end{tabular}
\end{table}  
\begin{table}[H]
\caption{Random Forest (RF)} 
\label{tbl:rf}
\centering 
\begin{tabular}{c c c c| c}
\toprule   
\thead{\textbf{}} & 
\thead{\textbf{Precision}} &
\thead{\textbf{Recall}} &
\thead{\textbf{F1-Score}} &
\thead{\textbf{Cases}} \\
  
\midrule
Yes & 0.65 & 0.35 & 0.46 &  201\\ 
\midrule
No &  0.94 & 0.98 & 0.96 & 2049  \\
\bottomrule
Macro Avg &  0.79 & 0.67 & 0.71 & 2250\\
\bottomrule
Weighted Avg & 0.91 & 0.92 & 0.91 & 2250

\end{tabular}
\end{table}  

\begin{table}[H]
\caption{Gradient Boosting (GB)} 
\label{tbl:gb}
\centering 
\begin{tabular}{c c c c| c}
\toprule   
\thead{\textbf{}} & 
\thead{\textbf{Precision}} &
\thead{\textbf{Recall}} &
\thead{\textbf{F1-Score}} &
\thead{\textbf{Cases}} \\
  
\midrule
Yes &0.71 & 0.37 & 0.49 &  201 \\ 
\midrule
No &   0.94 & 0.99 & 0.96 & 2049   \\
\bottomrule
Macro Avg & 0.83 & 0.68 & 0.73 & 2250 \\
\bottomrule
Weighted Avg & 0.92 & 0.93 & 0.92 & 2250

\end{tabular}
\end{table}

The precision, recall, F1-score, and cases results are all shown in the tables above. The Gradient Boosting Classifier showed the most accurate results out of all the other models. Although the F1-score of the answer choice "Yes" (0.49) of the Gradient Boosting Classifier is one of the highest models in the F1-score, it does not show a high enough value. Possible reasons for the low F1-score is that the NYTS data was insufficient and limited to create an accurate model. In addition, the ML models might not have been the best choices and there might be better algorithms to fit the NYTS data. Finally, the questions, or x variables, given and chosen might not be the most fitting ones and there are better options. Potential questions, or x variables, that could have been added to the NYTS data are detailed family history, respondent's location, and possibly the personality or qualities of the respondent. 

\begin{figure}[H]
\centering
\includegraphics[width=\textwidth]{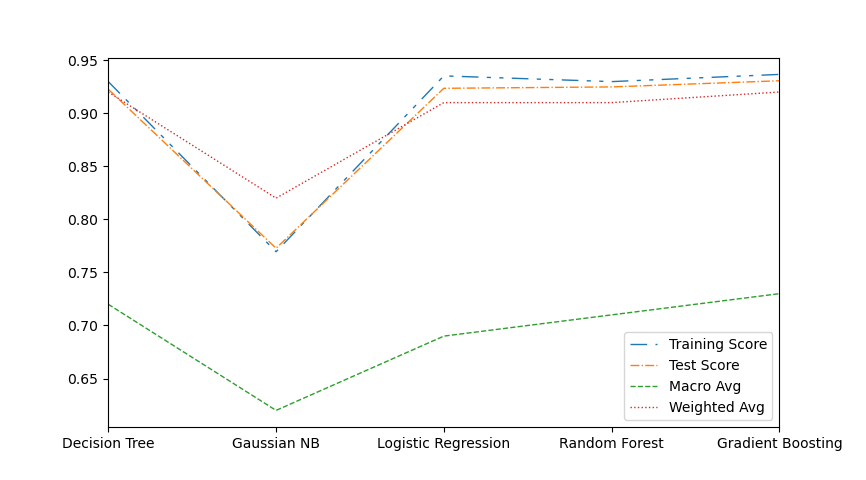}
\caption{Accuracy Results for the Five Algorithms}
\label{fig:resultforall}
\end{figure}

Table \ref{tbl:acml} and Figure \ref{fig:resultforall} shows the accuracy of training score and testing score of each machine learning model. 

\begin{table}[H]
\caption{Accuracy Results for ML Models in terms of Training and Test} 
\label{tbl:acml} 
\centering 
\begin{tabular}{c c c c c c}
\toprule 
\thead{\textbf{}} & 
\thead{\textbf{Decision Tree}} &
\thead{\textbf{Gaussian NB}} &
\thead{\textbf{Logistic Regression}} &
\thead{\textbf{Random Forest}} &
\thead{\textbf{Gradient Boosting}}\\
\midrule
Training Score & 0.9298 & 0.7695 & 0.9352 & 0.9298 & \textbf{0.9366}  \\ 
\midrule
Test Score & 0.9226 & 0.7728 & 0.9235 & 0.9248  & \textbf{0.9306} \\
\end{tabular}
\end{table}


The training scores resulted from  Cross Validation (CV). The training scores for Decision Tree, Gaussian NB, Logistic Regression, Random Forest, and Gradient Boosting were 0.9298, 0.7695, 0.9352, 0.9298, and 0.9366, respectively. Also, the test scores were 0.9226, 0.7728, 0.9235, 0.9248, and 0.9306, respectively.

Figure \ref{fig:resultforall} has the x-axis represented as the ML models and the y-axis represented as the training score, test score, macro average, and weighted average. The figure shows that Gradient Boosting classifier has the highest accuracy among others.

\subsection{Deployment}
\unskip

Our deployment process is through a public web page in which users (especially adolescents) enter their information for predicting their possibility of smoking cigarettes. The public web page (\href{http://nyts.pythonanywhere.com}{http://nyts.pythonanywhere.com}) gives the users instant access and quick response for the predicted result. The architecture and structure of the web page is introduced in Section 4.

\section{Prediction System for Future Smoking}

The best prediction model (i.e., Gradient Boosting) introduced in Section 3 is utilized in the public web page (\href{http://nyts.pythonanywhere.com}{http://nyts.pythonanywhere.com}) on the purpose of anti-smoking campaign for teenagers. This public web aims to inform teenagers of the possibility of future smoking, so that it may be helpful to prevent bad consequences of adolescents' health. In this section, we introduce the public web we developed. 

\begin{figure}[H]
\centering
\includegraphics[width=0.9\textwidth]{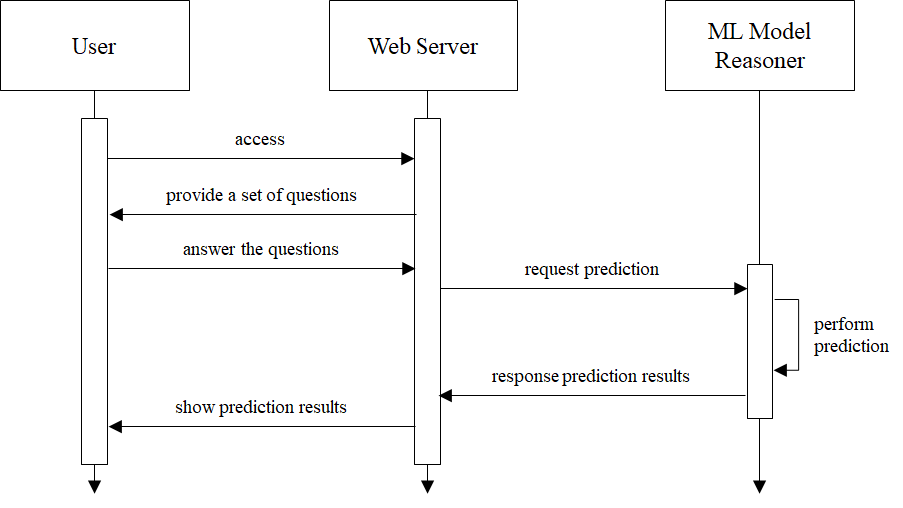}
\caption{Website Sequence Diagram}
\label{fig:wsos}
\end{figure}

Figure \ref{fig:wsos} shows a sequence diagram representing about how the user (e.g., e-cigarette or non-smoking teenagers) interacts with the website. The user, web server, and ML model reasoner represented by the boxes denotes main entities. The ML model reasoner contains the Gradient Boosting model that was learned in Section 3 using the NYTS data. The path for each entity goes from top to bottom and the vertical rectangle means the life cycle of the entity. The arrow lines across the vertical rectangles represent event flows. The following list summarize the event flows.
 
\begin{enumerate}
  \item A user enters the web server.
  \item The web server shows 47 questions to the user.
  \item The user inputs all answers for the questions.
  \item The web server receives the answers from the user and sends them to the ML model reasoner.
  \item The ML model reasoner performs prediction using the answers and sends the results back to the web server.
  \item The web server displays the prediction results using a chart.
\end{enumerate} 

The web server was developed in the Python environment. We used Flask v1.1.1, a Python-based web server, and "pythonanywhere.com" for a web hosting. The Gradient Boosting model was learned using Sciki-Learn v0.21.3 and was stored using Pickle python module to serialize and de-serialize for the learned model. For reasoning of the model, again Sciki-Learn v0.21.3 was used. The web page example can be found in Figure \ref{fig:wsex}.

\section{Conclusion}

E-cigarette use has increased among adolescents. This is a worldwide problem, because it has been stated in many researches mentioned in the introduction that e-cigarette use can cause future use of cigarettes. Since e-cigarette is a recent rising issue, there is little research done on this topic, compared to smoking cigarettes. Even among the researches done, there is a lack of researches implementing prediction models, which are more practical in preventing adolescents from using (e-)cigarettes. Thus, we researched using the 2018 NYTS data and developed multiple prediction models to predict a adolescent's intention to smoke cigarette. 

The most accurate prediction model was Gradient Boosting Classifier with an overall accuracy of 93\%. This model was applied in the website we designed to allow the public to input their information in respect to tobacco products, including e-cigarette, cigarette, and cigar. With this information, the algorithm can predict the respondee's probability of future of smoking. This will help the public become more aware about certain factors in their lives and be attentive about their drug use or how their environment can affect their intention to smoke cigarettes. 

Further research could include a wider range of ages, since our research is mainly focused on adolescents rather than adults. In order to improve the accuracy of the prediction model, it is essential to increase the amount of data or choose better, more fitting, variables. 
 


 

\appendixtitles{no} 
\appendix
\section{}
\unskip
\subsection{}
These are the list of questions that were used for data analysis and what they were used for. The "Number" column specifies the question number from the 2018 NYTS questionnaire. The "Question" column lists out the exact question used from the questionnaire and "Where to Use" column specifies the purpose of using the question. The term "Predictor Variable" means that the question was considered a factor, or x variable, in predicting the intention for a person to smoke. "Data Selection for Non-Smoker" means that the question was utilized to separate the data into non-smoker and smoker groups. "Data Selection for Smoking Intention" means that the question was used as a target question, or a y variable, because the goal is to predict intention to smoke cigarette. "Data Selection for Non-E-Smoker" means that the question was used to separate the non-e-cigarette smokers from the e-cigarette smokers.

\begin{longtable}{c c c}

\toprule
\textbf{Number} & \textbf{Question} & \textbf{Where to Use} \\
\midrule
\endhead
 
\midrule
Q1 & How old are you? &  Predictor Variable \\ 
Q2 & What is your sex? & Predictor Variable \\ 
Q3 & What grade are you in? & Predictor Variable \\  
Q4 & Are you Hispanic, Latino, Latina, or of Spanish origin? & Predictor Variable \\  
Q5 & What race or races do you consider yourself to be? & Predictor Variable \\  
Q6 &  Have you ever been curious about smoking a cigarette? & Predictor Variable \\  
Q7 & Have you ever tried cigarette smoking, even one or two puffs? & \thead{Data Selection \\ for Non-Smoker}\\  
Q15 & Do you think that you will try a cigarette soon? & \thead{Data Selection \\ for Smoking Intention} \\  
Q16 & Do you think you will smoke a cigarette in the next year? & \thead{Data Selection \\ for Smoking Intention} \\  
Q17 & \thead{If one of your best friends were to offer you a cigarette,\\  would you smoke it?} & \thead{Data Selection \\ for Smoking Intention} \\  
Q18 & \thead{Have you ever been curious about smoking a cigar, cigarillo,\\  or little cigar?} & Predictor Variable \\  
Q19 & \thead{Have you ever tried smoking cigars, cigarillos, or \\ little cigars even one or two puffs?} & \thead{Data Selection \\ for Non-Smoker} \\  
Q23 & \thead{Have you ever been curious about using chewing tobacco, snuff, \\ or dip?} & Predictor Variable \\  
Q24 & \thead{Have you ever used chewing tobacco, snuff, or dip,\\  such as Redman, Levi Garrett, Beechnut, Skoal, Skoal Bandits, or Copenhagen, \\ even just a small amount?} & \thead{Data Selection \\ for Non-Smoker} \\ 
Q27 & \thead{Have you ever been curious about using an e-cigarette?} & Predictor Variable \\  
Q28 & \thead{Have you ever used an e-cigarette, even once or twice?} & \thead{Data Selection \\ for Non-E-Smoker} \\  
Q29 & \thead{How old were you when you first tried using an e-cigarette, \\ even once or twice?} & Predictor Variable \\  
Q30 & \thead{In total, on how many days have you used e-cigarettes \\ in your entire life?} & Predictor Variable \\  
Q31 & \thead{During the past 30 days, on how many days did you use e-cigarettes?} & Predictor Variable \\  
Q32 & \thead{During the past 30 days, where did you get or buy \\ the e-cigarettes that you have used?} & Predictor Variable \\  
Q33 & \thead{What are the reasons you have used e-cigarettes?} & Predictor Variable \\  
Q34 & \thead{Have you ever used marijuana, marijuana concentrates, \\ marijuana waxes, THC, or hash oils in an e-cigarette?} & Predictor Variable \\  
Q35 & \thead{Do you think that you will try an e-cigarette soon?} & Predictor Variable \\  
Q36 & \thead{Do you think you will use an e-cigarette in the next year?} & Predictor Variable \\  
Q37 & \thead{If one of your best friends were to offer you an e-cigarette,\\  would you use it?} & Predictor Variable \\  
Q38 & \thead{Have you ever been curious about smoking tobacco in a hookah or waterpipe?} & Predictor Variable \\  
Q39 & \thead{Have you ever tried smoking tobacco in a hookah or \\ waterpipe, even one or two puffs?} & \thead{Data Selection \\ for Non-Smoker} \\  
Q43 & \thead{Do you think that you will try smoking tobacco in \\ a hookah or waterpipe soon?} & \thead{Data Selection \\ for Smoking Intention} \\  
Q44 & \thead{Do you think you will smoke tobacco in a hookah or\\  waterpipe in the next year?} & \thead{Data Selection \\ for Smoking Intention} \\  
Q45 & \thead{If one of your best friends were to offer you a hookah or \\ waterpipe with tobacco, would you try it?} & \thead{Data Selection \\ for Smoking Intention} \\  
Q59 & \thead{During the past 30 days, did anyone refuse to sell you \\ any tobacco products because of your age?} & \thead{Data Selection \\ for Non-Smoker} \\  
Q61 & \thead{During the past 30 days, how often did you see a warning label \\ on a cigar, cigarillo, or little cigar package?} & Predictor Variable \\  
Q62 & \thead{During the past 30 days, how often did you see a warning label \\ on an e-cigarette package?} & Predictor Variable \\  
Q63 & \thead{During the past 30 days, how often did you see a warning label \\ on a package of hookah tobacco?} & Predictor Variable \\  
Q64 & \thead{In the past 12 months, have you seen or heard The Real Cost, \\ on television, the internet, social media, or radio as part of ads about tobacco?} & Predictor Variable \\  
Q65 & \thead{How much do you think people harm themselves when they smoke \\  cigarettes some days but not every day?} & Predictor Variable \\  
Q66 & \thead{How much do you think people harm themselves when they use \\ chewing tobacco, snuff, dip, or snus, some days but not every day?} & Predictor Variable \\  
Q67 & \thead{Do you believe that chewing tobacco, snuff, dip, or snus is \\ (LESS ADDICTIVE, EQUALLY ADDICTIVE, or MORE ADDICTIVE) \\ than cigarettes?} & Predictor Variable \\  
Q68 & \thead{How much do you think people harm themselves when they use \\  e-cigarettes some days but not every day?} & Predictor Variable \\  
Q69 & \thead{Do you believe that e-cigarettes are (LESS ADDICTIVE, \\ EQUALLY ADDICTIVE, or MORE ADDICTIVE) than cigarettes?} & Predictor Variable \\  
Q70 & \thead{How much do you think people harm themselves when they \\ smoke tobacco in a hookah or waterpipe some days but not every day?} & Predictor Variable \\  
Q71 & \thead{Do you believe that smoking tobacco in a hookah or waterpipe \\ is (LESS ADDICTIVE, EQUALLY ADDICTIVE, or MORE ADDICTIVE) \\ than cigarettes?} & Predictor Variable \\  
Q72 & \thead{How strongly do you agree with the statement ‘All tobacco products \\ are dangerous’?} & Predictor Variable \\  
Q73 & \thead{Not including the vapor from e-cigarettes, do you think \\ that breathing smoke from other people’s cigarettes or other tobacco products causes} & Predictor Variable \\  
Q74 & \thead{When you are using the Internet, how often do you see ads \\ or promotions for cigarettes or other tobacco products?} & Predictor Variable \\  
Q75 & \thead{When you read newspapers or magazines, how often do you see ads \\ or promotions for cigarettes or other tobacco products?} & Predictor Variable \\  
Q76 & \thead{When you go to a convenience store, supermarket, or \\ gas station, how often do you see ads or promotions \\for cigarettes or other tobacco products?} & Predictor Variable \\  
Q77 & \thead{When you watch TV or go to the movies, how often do you see \\ ads or promotions for cigarettes or other tobacco products?} & Predictor Variable \\  
Q78 & \thead{When you are using the Internet, how often do you see ads \\ or promotions for e-cigarettes?} & Predictor Variable \\  
Q79 & \thead{When you read newspapers or magazines, how often do you see ads \\ or promotions for e-cigarettes?} & Predictor Variable \\  
Q80 & \thead{When you go to a convenience store, supermarket, or gas station, \\  how often do you see ads or promotions for e-cigarettes?} & Predictor Variable \\  
Q81 & \thead{When you watch TV, how often do you see ads or promotions for e-cigarettes?} & Predictor Variable \\  
Q82 & \thead{During the past 7 days, on how many days did someone smoke \\ tobacco products in your home while you were there?} & Predictor Variable \\  
Q83 & \thead{During the past 7 days, on how many days did you ride in a vehicle \\ when someone was smoking a tobacco product?} & Predictor Variable \\  
Q84 & \thead{During the past 30 days, on how many days did you breathe \\ the smoke from someone who was smoking tobacco products \\ in an indoor or outdoor public place?} & Predictor Variable \\  
Q85 & \thead{During the past 30 days, on how many days did you breathe \\ the vapor from someone who was using an e-cigarette \\ in an indoor or outdoor public place?} & Predictor Variable \\  
Q86 & \thead{Does anyone who lives with you now…?} & Predictor Variable \\  
Q87 & \thead{Do you speak a language other than English at home?} & Predictor Variable \\  
Q88 & \thead{Because of a physical, mental, or emotional condition, \\ do you have serious difficulty concentrating,remembering, or making decisions?} & Predictor Variable \\  
 
\midrule  
\label{tb:quest88}
\end{longtable} 

\subsection{} 
This is a glimpse of the website. After answering total of 47 questions and clicking submit, the bubble will fill up and show the probability of a user to smoker in the future.

\begin{figure}[H]
\centering
\includegraphics[width=0.5\textwidth]{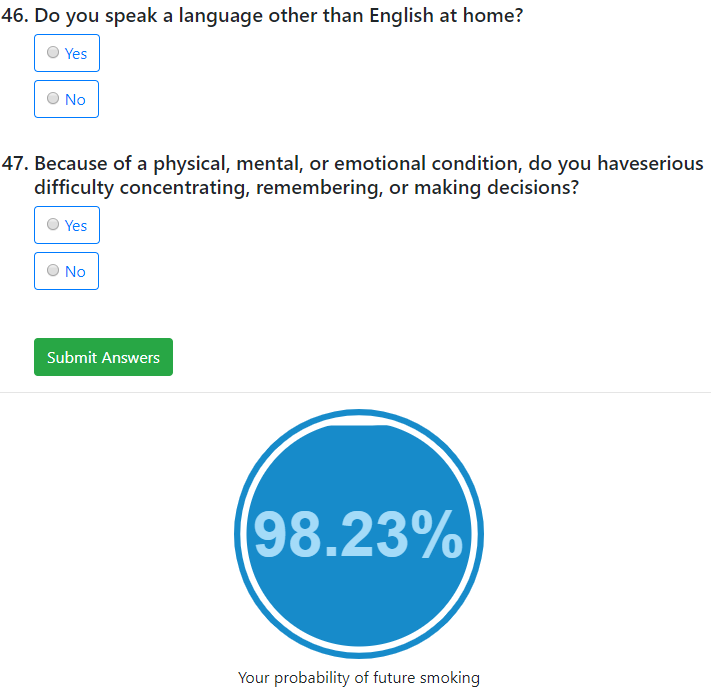}
\caption{Website Example}
\label{fig:wsex}
\end{figure}

\reftitle{References}




\externalbibliography{yes}
\bibliography{my.bib}
  

\end{document}